\def\year{2017}\relax
\documentclass[letterpaper]{article}
\usepackage{aaai17}
\usepackage{times}
\usepackage{helvet}
\usepackage{courier}
\usepackage{graphicx}

\usepackage{amssymb}
\usepackage{amsmath}
\usepackage{booktabs}
\usepackage{color}

\usepackage{hyperref}
\hypersetup{
    colorlinks=true,
    citecolor= black,
    linkcolor=blue,
    filecolor=blue,      
    urlcolor=blue,
}

\newcommand{\newcite}[1]{\citeauthor{#1} (\citeyear{#1})}

\begin{document}
%
\title{Playing FPS Games with Deep Reinforcement Learning}
\author{Guillaume Lample\thanks{The authors contributed equally to this work.}, Devendra Singh Chaplot\footnotemark[1]
\\ \{glample,chaplot\}@cs.cmu.edu \\ School of Computer Science \\ Carnegie Mellon University}
\maketitle
\begin{abstract}
Advances in deep reinforcement learning have allowed autonomous agents to perform well on Atari games, often outperforming humans, using only raw pixels to make their decisions. However, most of these games take place in 2D environments that are fully observable to the agent. In this paper, we present the first architecture to tackle 3D environments in first-person shooter games, that involve partially observable states. Typically, deep reinforcement learning methods only utilize visual input for training. We present a method to augment these models to exploit game feature information such as the presence of enemies or items, during the training phase. Our model is trained to simultaneously learn these features along with minimizing a Q-learning objective, which is shown to dramatically improve the training speed and performance of our agent. Our architecture is also modularized to allow different models to be independently trained for different phases of the game. We show that the proposed architecture substantially outperforms built-in AI agents of the game as well as average humans in deathmatch scenarios.
\end{abstract}

\section{Introduction}
Deep reinforcement learning has proved to be very successful in mastering human-level control policies in a wide variety of tasks such as object recognition with visual attention \cite{ba2014multiple}, high-dimensional robot control \cite{levine2016end} and solving physics-based control problems \cite{heess2015learning}. In particular, Deep Q-Networks (DQN) are shown to be effective in playing Atari 2600 games \cite{mnih2013playing} and more recently, in defeating world-class Go players \cite{silver2016mastering}.

However, there is a limitation in all of the above applications in their assumption of having the full knowledge of the current state of the environment, which is usually not true in real-world scenarios. In the case of partially observable states, the learning agent needs to remember previous states in order to select optimal actions. Recently, there have been attempts to handle partially observable states in deep reinforcement learning by introducing recurrency in Deep Q-networks. For example, \newcite{hausknecht2015deep} use a deep recurrent neural network, particularly a Long-Short-Term-Memory (LSTM) Network, to learn the Q-function to play Atari 2600 games. \newcite{foerster2016learning} consider a multi-agent scenario where they use deep distributed recurrent neural networks to communicate between different agent in order to solve riddles. The use of recurrent neural networks is effective in scenarios with partially observable states due to its ability to remember information for an arbitrarily long amount of time.

\begin{figure}[h]
\centering
\includegraphics[width=1.0\linewidth,height=\textheight,keepaspectratio]{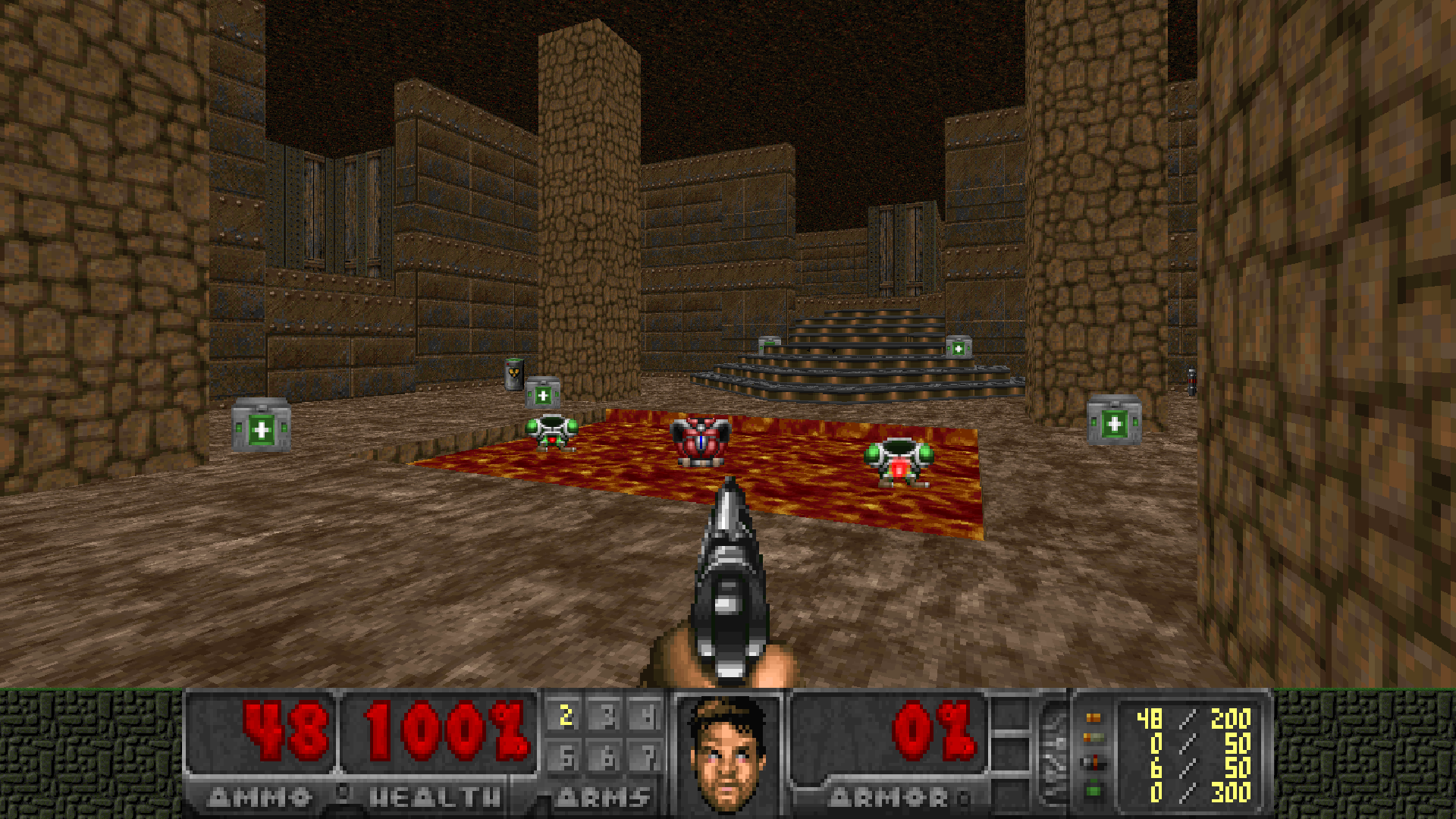}
\caption{A screenshot of Doom.}
\label{fig:scrn}
\end{figure}

Previous methods have usually been applied to 2D environments that hardly resemble the real world. In this paper, we tackle the task of playing a First-Person-Shooting (FPS) game in a 3D environment. This task is much more challenging than playing most Atari games as it involves a wide variety of skills, such as navigating through a map, collecting items, recognizing and fighting enemies, etc. Furthermore, states are partially observable, and the agent navigates a 3D environment in a first-person perspective, which makes the task more suitable for real-world robotics applications.

In this paper, we present an AI-agent\footnote{Code: \url{https://github.com/glample/Arnold}} for playing deathmatches\footnote{A deathmatch is a scenario in FPS games where the objective is to maximize the number of kills by a player/agent.} in FPS games using only the pixels on the screen. Our agent divides the problem into two phases: navigation (exploring the map to collect items and find enemies) and action (fighting enemies when they are observed), and uses separate networks for each phase of the game. Furthermore, the agent infers high-level game information, such as the presence of enemies on the screen, to decide its current phase and to improve its performance. We also introduce a method for co-training a DQN with game features, which turned out to be critical in guiding the convolutional layers of the network to detect enemies. We show that co-training significantly improves the training speed and performance of the model.

We evaluate our model on the two different tasks adapted from the Visual Doom AI Competition (ViZDoom)\footnote{ViZDoom Competition at IEEE Computational Intelligence And Games (CIG) Conference, 2016 (http://vizdoom.cs.put.edu.pl/competition-cig-2016)} using the API developed by \newcite{kempka2016vizdoom} (Figure~\ref{fig:scrn} shows a screenshot of Doom).
The API gives a direct access to the Doom game engine and allows to synchronously send commands to the game agent and receive inputs of the current state of the game. We show that the proposed architecture substantially outperforms built-in AI agents of the game as well as humans in deathmatch scenarios and we demonstrate the importance of each component of our architecture.

\section{Background}
\label{sec:background}

Below we give a brief summary of the DQN and DRQN models.

\subsection{Deep Q-Networks}

Reinforcement learning deals with learning a policy for an agent interacting in an unknown environment. At each step, an agent observes the current state $s_t$ of the environment, decides of an action $a_t$ according to a policy $\pi$, and observes a reward signal $r_t$. The goal of the agent is to find a policy that maximizes the expected sum of discounted rewards $R_t$

$$ R_t = \sum\limits_{t'=t}^T \gamma^{t'-t} r_{t'}$$

\noindent where T is the time at which the game terminates, and $\gamma \in \left[ 0, 1\right]$ is a discount factor that determines the importance of future rewards. The $Q$-function of a given policy $\pi$ is defined as the expected return from executing an action $a$ in a state $s$:

$$Q^{\pi}(s, a) = \mathbb{E} \left[ R_t | s_t = s, a_t = a \right]$$

\noindent It is common to use a function approximator to estimate the action-value function $Q$. In particular, DQN uses a neural network parametrized by $\theta$, and the idea is to obtain an estimate of the $Q$-function of the current policy which is close to the optimal $Q$-function $Q^{*}$ defined as the highest return we can expect to achieve by following any strategy:

$$Q^{*}(s, a) = \max_{\pi} \mathbb{E} \left[ R_t | s_t = s, a_t = a \right] = \max_{\pi} Q^{\pi}(s, a)$$

\noindent In other words, the goal is to find $\theta$ such that $Q_{\theta}(s, a) \approx Q^{*}(s, a)$. The optimal $Q$-function verifies the Bellman optimality equation

$$Q^{*}(s, a) = \mathbb{E} \big[ r + \gamma \max_{a'} Q^{*}(s', a') | s, a \big]$$

\noindent If $Q_{\theta} \approx Q^{*}$, it is natural to think that $Q_{\theta}$ should be close from also verifying the Bellman equation. This leads to the following loss function:

$$L_t (\theta_t) = \mathbb{E}_{s,a,r,s'} \bigg[ \big( y_t - Q_{\theta_t}(s, a) \big)^{2} \bigg]$$

\noindent where $t$ is the current time step, and $y_t = r + \gamma \max_{a'} Q_{\theta_t}(s', a')$. The value of $y_t$ is fixed, which leads to the following gradient:

$$\nabla_{\theta_t} L_t (\theta_t) = \mathbb{E}_{s,a,r,s'} \bigg[ \big(y_t - Q_{\theta}(s, a) \big) \nabla_{\theta_t} Q_{\theta_t}(s, a) \bigg]$$

\noindent Instead of using an accurate estimate of the above gradient, we compute it using the following approximation:

$$\nabla_{\theta_t} L_t (\theta_t) \approx \big(y_t - Q_{\theta}(s, a) \big) \nabla_{\theta_t} Q_{\theta_t}(s, a)$$

Although being a very rough approximation, these updates have been shown to be stable and to perform well in practice.

Instead of performing the Q-learning updates in an online fashion, it is popular to use experience replay \cite{lin1993reinforcement} to break correlation between successive samples. At each time steps, agent experiences $(s_t, a_t, r_t, s_{t + 1})$ are stored in a replay memory, and the Q-learning updates are done on batches of experiences randomly sampled from the memory.

At every training step, the next action is generated using an $\epsilon$-greedy strategy: with a probability $\epsilon$ the next action is selected randomly, and with probability $1 - \epsilon$ according to the network best action. In practice, it is common to start with $\epsilon = 1$ and to progressively decay $\epsilon$.

\subsection{Deep Recurrent Q-Networks}

The above model assumes that at each step, the agent receives a full observation $s_t$ of the environment - as opposed to games like Go, Atari games actually rarely return a full observation, since they still contain hidden variables, but the current screen buffer is usually enough to infer a very good sequence of actions. But in partially observable environments, the agent only receives an observation $o_t$ of the environment which is usually not enough to infer the full state of the system. A FPS game like DOOM, where the agent field of view is limited to 90° centered around its position, obviously falls into this category.

To deal with such environments, \newcite{hausknecht2015deep} introduced the Deep Recurrent Q-Networks (DRQN), which does not estimate $Q(s_t, a_t)$, but $Q(o_t, h_{t - 1}, a_t)$, where $h_t$ is an extra input returned by the network at the previous step, that represents the hidden state of the agent. A recurrent neural network like a LSTM \cite{hochreiter1997long} can be implemented on top of the normal DQN model to do that. In that case, $h_t = \textrm{LSTM}(h_{t - 1}, o_t)$, and we estimate $Q(h_t, a_t)$. Our model is built on top of the DRQN architecture.

\begin{figure*}
\includegraphics[width=\textwidth,height=\textheight,keepaspectratio]{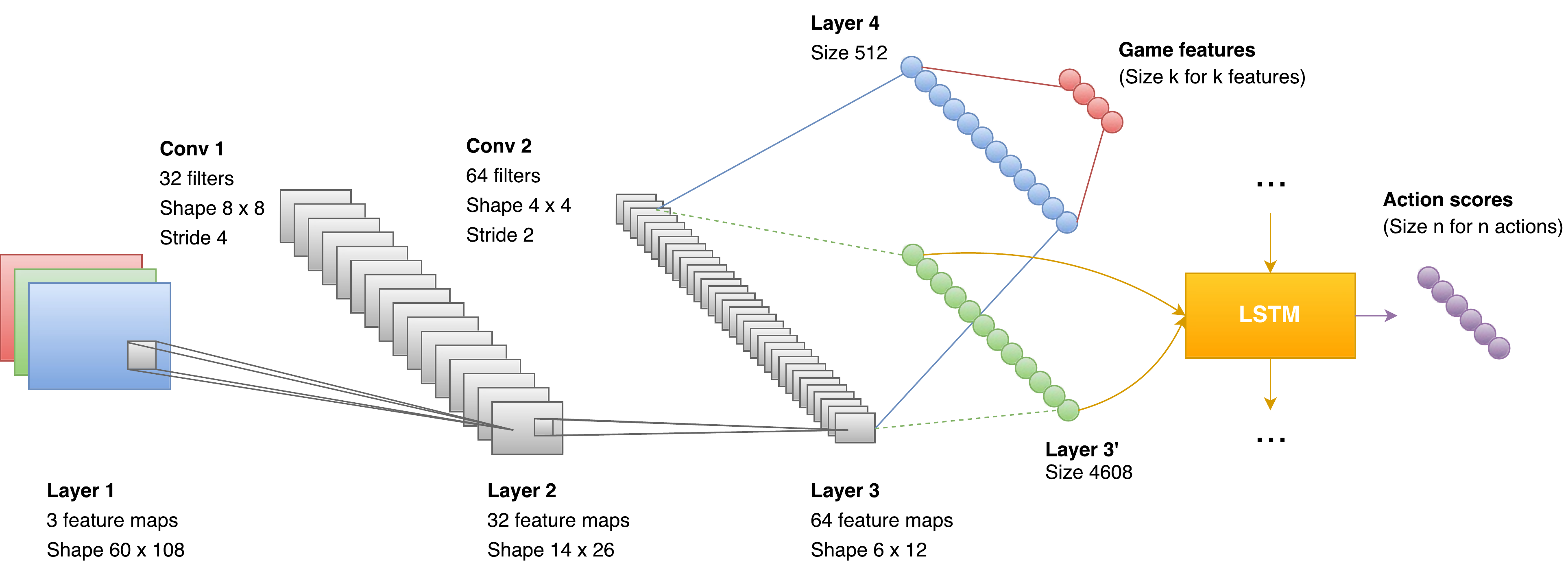}
\caption{An illustration of the architecture of our model. The input image is given to two convolutional layers. The output of the convolutional layers is split into two streams. The first one (bottom) flattens the output (layer 3') and feeds it to a LSTM, as in the DRQN model. The second one (top) projects it to an extra hidden layer (layer 4), then to a final layer representing each game feature. During the training, the game features and the Q-learning objectives are trained jointly.}
\label{fig:model-architecture}
\end{figure*}

\section{Model}
\label{sec:model}
Our first approach to solving the problem was to use a baseline DRQN model. Although this model achieved good performance in relatively simple scenarios (where the only available actions were to turn or attack), it did not perform well on deathmatch tasks. The resulting agents were firing at will, hoping for an enemy to come under their lines of fire. Giving a penalty for using ammo did not help: with a small penalty, agents would keep firing, and with a big one they would just never fire.

\subsection{Game feature augmentation}
\label{sec:game-feature-augmentation}

We reason that the agents were not able to accurately detect enemies. The ViZDoom environment gives access to internal variables generated by the game engine. We modified the game engine so that it returns, with every frame, information about the visible entities. Therefore, at each step, the network receives a frame, as well as a Boolean value for each entity, indicating whether this entity appears in the frame or not (an entity can be an enemy, a health pack, a weapon, ammo, etc). Although this internal information is not available at test time, it can be exploited during training. We modified the DRQN architecture to incorporate this information and to make it sensitive to game features. In the initial model, the output of the convolutional neural network (CNN) is given to a LSTM that predicts a score for each action based on the current frame and its hidden state. We added two fully-connected layers of size $512$ and $k$ connected to the output of the CNN, where $k$ is the number of game features we want to detect. At training time, the cost of the network is a combination of the normal DRQN cost and the cross-entropy loss. Note that the LSTM only takes as input the CNN output, and is never directly provided with the game features. An illustration of the architecture is presented in Figure~\ref{fig:model-architecture}.

\begin{figure*}
\minipage{0.32\textwidth}
\includegraphics[width=\linewidth,height=\textheight,keepaspectratio]{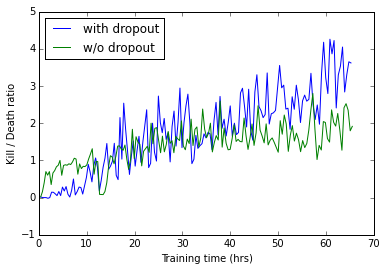}
\endminipage\hfill
\minipage{0.33\textwidth}
\includegraphics[width=\linewidth,height=\textheight,keepaspectratio]{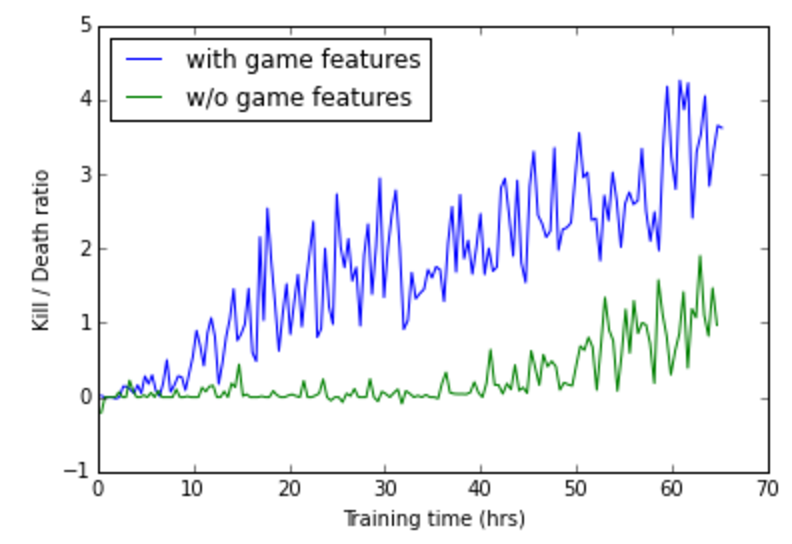}
\endminipage\hfill
\minipage{0.32\textwidth}%
\includegraphics[width=\linewidth,height=\textheight,keepaspectratio]{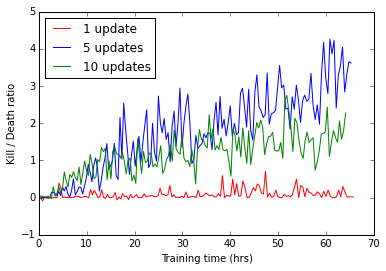}
\endminipage
\caption{Plot of K/D score of action network on limited deathmatch as a function of training time (a) with and without dropout (b) with and without game features, and (c) with different number of updates in the LSTM.}
\label{fig:graphs}
\end{figure*}

Although a lot of game information was available, we only used an indicator about the presence of enemies on the current frame. Adding this game feature dramatically improved the performance of the model on every scenario we tried. Figure~\ref{fig:graphs} shows the performance of the DRQN with and without the game features. We explored other architectures to incorporate game features, such as using a separate network to make predictions and reinjecting the predicted features into the LSTM, but this did not achieve results better than the initial baseline, suggesting that sharing the convolutional layers is decisive in the performance of the model. Jointly training the DRQN model and the game feature detection allows the kernels of the convolutional layers to capture the relevant information about the game. In our experiments, it only takes a few hours for the model to reach an optimal enemy detection accuracy of 90\%. After that, the LSTM is given features that often contain information about the presence of enemy and their positions, resulting in accelerated training.

Augmenting a DRQN model with game features is straightforward. However, the above method can not be applied easily to a DQN model. Indeed, the important aspect of the model is the sharing of the convolution filters between predicting game features and the Q-learning objective. The DRQN is perfectly adapted to this setting since the network takes as input a single frame, and has to predict what is visible in this specific frame. However, in a DQN model, the network receives $k$ frames at each time step, and will have to predict whether some features appear in the last frame only, independently of the content of the $k-1$ previous frames. Convolutional layers do not perform well in this setting, and even with dropout we never obtained an enemy detection accuracy above 70\% using that model.

\subsection{Divide and conquer} 
The deathmatch task is typically divided into two phases, one involves exploring the map to collect items and to find enemies, and the other consists in fighting enemies \cite{mcpartland2008learning,tastan2011learning}. We call these phases the navigation and action phases. Having two networks work together, each trained to act in a specific phase of the game should naturally lead to a better overall performance. Current DQN models do not allow for the combination of different networks optimized on different tasks. However, the current phase of the game can be determined by predicting whether an enemy is visible in the current frame (action phase) or not (navigation phase), which can be inferred directly from the game features present in the proposed model architecture.

There are various advantages of splitting the task into two phases and training a different network for each phase. First, this makes the architecture modular and allows different models to be trained and tested independently for each phase. Both networks can be trained in parallel, which makes the training much faster as compared to training a single network for the whole task. Furthermore, the navigation phase only requires three actions (move forward, turn left and turn right), which dramatically reduces the number of state-action pairs required to learn the Q-function, and makes the training much faster \cite{gaskett1999q}. More importantly, using two networks also mitigates ``camper'' behavior, i.e. tendency to stay in one area of the map and wait for enemies, which was exhibited by the agent when we tried to train a single DQN or DRQN for the deathmatch task. 

We trained two different networks for our agent. We used a DRQN augmented with game features for the action network, and a simple DQN for the navigation network. During the evaluation, the action network is called at each step. If no enemies are detected in the current frame, or if the agent does not have any ammo left, the navigation network is called to decide the next action. Otherwise, the decision is given to the action network. Results in Table~\ref{tab:splitting} demonstrate the effectiveness of the navigation network in improving the performance of our agent.

\begin{figure}
\includegraphics[width=\linewidth,height=\textheight,keepaspectratio]{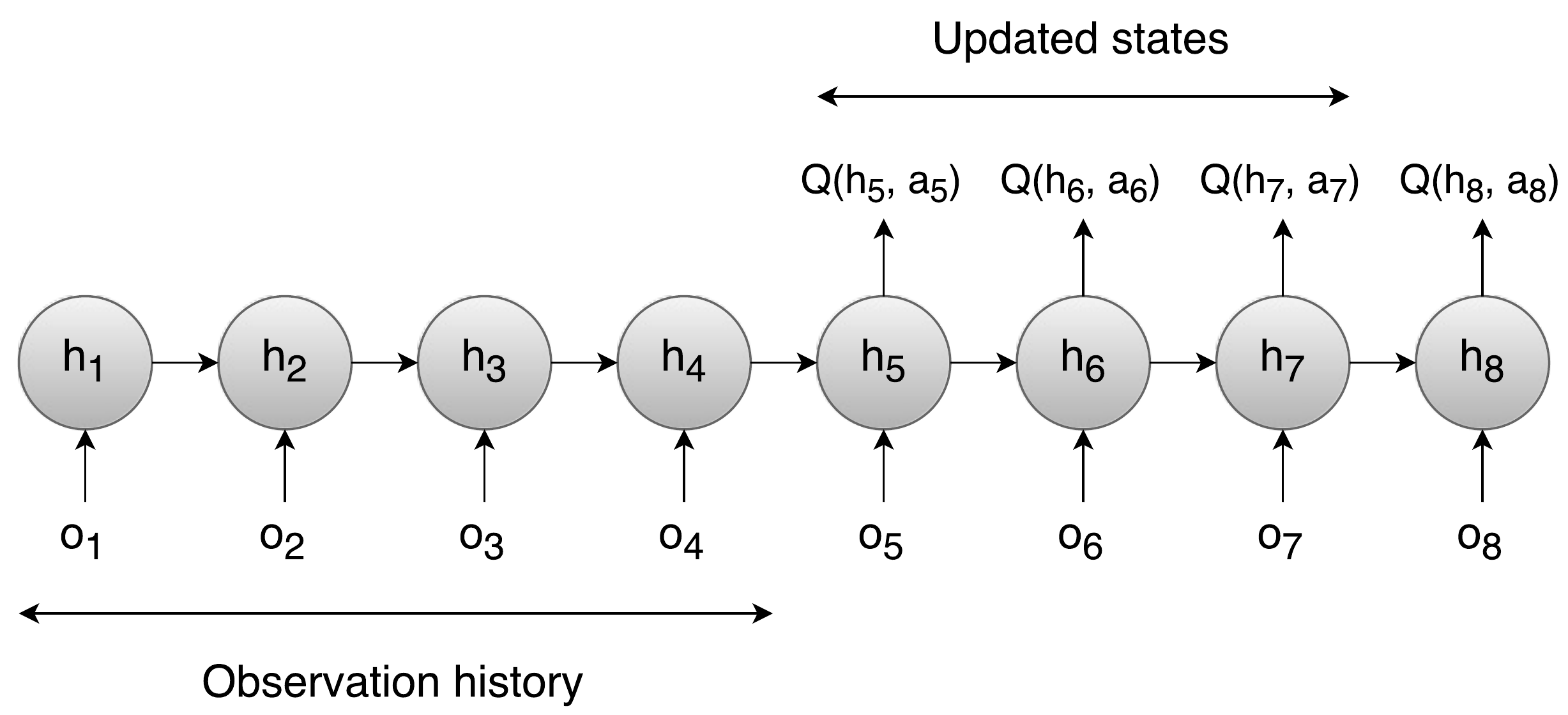}
\caption{DQN updates in the LSTM. Only the scores of the actions taken in states 5, 6 and 7 will be updated. First four states provide a more accurate hidden state to the LSTM, while the last state provide a target for state 7.}
\label{fig:drqn-updates}
\end{figure}

\section{Training}
\label{sec:training}
\subsection{Reward shaping}
The score in the deathmatch scenario is defined as the number of frags, i.e. number of kills minus number of suicides. If the reward is only based on the score, the replay table is extremely sparse w.r.t state-action pairs having non-zero rewards, which makes it very difficult for the agent to learn favorable actions. Moreover, rewards are extremely delayed and are usually not the result of a specific action: getting a positive reward requires the agent to explore the map to find an enemy and accurately aim and shoot it with a slow projectile rocket. The delay in reward makes it difficult for the agent to learn which set of actions is responsible for what reward.
To tackle the problem of sparse replay table and delayed rewards, we introduce reward shaping, i.e. the modification of reward function to include small intermediate rewards to speed up the learning process \cite{ng2003shaping}. In addition to positive reward for kills and negative rewards for suicides, we introduce the following intermediate rewards for shaping the reward function of the action network:
\begin{itemize}
\item positive reward for object pickup (health, weapons and ammo)
\item negative reward for loosing health (attacked by enemies or walking on lava)
\item negative reward for shooting, or loosing ammo
\end{itemize}

We used different rewards for the navigation network. Since it evolves on a map without enemies and its goal is just to gather items, we simply give it a positive reward when it picks up an item, and a negative reward when it's walking on lava. We also found it very helpful to give the network a small positive reward proportional to the distance it travelled since the last step. That way, the agent is faster to explore the map, and avoids turning in circles.

\subsection{Frame skip}
Like in most previous approaches, we used the frame-skip technique \cite{bellemare2012arcade}. In this approach, the agent only receives a screen input every $k + 1$ frames, where $k$ is the number of frames skipped between each step. The action decided by the network is then repeated over all the skipped frames. A higher frame-skip rate accelerates the training, but can hurt the performance. Typically, aiming at an enemy sometimes requires to rotate by a few degrees, which is impossible when the frame skip rate is too high, even for human players, because the agent will repeat the rotate action many times and ultimately rotate more than it intended to. A frame skip of $k = 4$ turned out to be the best trade-off.

\subsection{Sequential updates}

To perform the DRQN updates, we use a different approach from the one presented by \newcite{hausknecht2015deep}. A sequence of $n$ observations $o_1, o_2, ..., o_n$ is randomly sampled from the replay memory, but instead of updating all action-states in the sequence, we only consider the ones that are provided with enough history. Indeed, the first states of the sequence will be estimated from an almost non-existent history (since $h_0$ is reinitialized at the beginning of the updates), and might be inaccurate. As a result, updating them might lead to imprecise updates.

To prevent this problem, errors from states $o_1 ... o_h$, where $h$ is the minimum history size for a state to be updated, are not backpropagated through the network. Errors from states $o_{h + 1} .. o_{n - 1}$ will be backpropagated, $o_n$ only being used to create a target for the $o_{n - 1}$ action-state. An illustration of the updating process is presented in Figure~\ref{fig:drqn-updates}, where $h=4$ and $n=8$. In all our experiments, we set the minimum history size to $4$, and we perform the updates on $5$ states.
Figure~\ref{fig:graphs} shows the importance of selecting an appropriate number of updates. Increasing the number of updates leads to high correlation in sampled frames, violating the DQN random sampling policy, while decreasing the number of updates makes it very difficult for the network to converge to a good policy.

\setlength\tabcolsep{4.8pt}
\begin{table}[]
\centering
\label{my-label}

\begin{tabular}{@{}lccccc@{}}
\toprule
                         & \multicolumn{2}{c}{Single Player} & \multicolumn{1}{l}{} & \multicolumn{2}{c}{Multiplayer} \\ \midrule
Evaluation Metric        & Human           & Agent          & \multicolumn{1}{l}{} & Human          & Agent         \\ \midrule
Number of objects        & 5.2             & 9.2             &                      & 6.1            & 10.5           \\
Number of kills          & 12.6            & 27.6            &                      & 5.5            & 8.0            \\
Number of deaths         & 8.3             & 5.0             &                      & 11.2           & 6.0            \\
Number of suicides       & 3.6             & 2.0             &                      & 3.2            & 0.5            \\
K/D Ratio      & 1.52            & 5.12            &                      & 0.49           & 1.33           \\ \bottomrule
\end{tabular}
\caption{Comparison of human players with agent. Single player scenario is both humans and the agent playing against bots in separate games. Multiplayer scenario is agent and human playing against each other in the same game.}
\label{tab:results}
\end{table}

\section{Experiments}
\label{sec:experiment}
\setlength\tabcolsep{8pt}
\begin{table*}[]
\centering
\begin{tabular}{@{}lcccccccc@{}}
\toprule
                         & \multicolumn{2}{c}{Limited Deathmatch}                                                                                    & \multicolumn{1}{l}{} & \multicolumn{5}{c}{Full Deathmatch}                                                                                                                                                                                                                                        \\ \midrule
                         & \multicolumn{2}{c}{Known Map}                                                                                             & \multicolumn{1}{l}{} & \multicolumn{2}{c}{Train maps}                                                                                 & \multicolumn{1}{l}{} & \multicolumn{2}{c}{Test maps}                                                                                  \\ \midrule
Evaluation Metric        & \begin{tabular}[c]{@{}c@{}}Without\\ navigation\end{tabular} & \begin{tabular}[c]{@{}c@{}}With \\ navigation\end{tabular} & \multicolumn{1}{l}{} & \begin{tabular}[c]{@{}c@{}}Without\\ navigation\end{tabular} & \begin{tabular}[c]{@{}c@{}}With\\ navigation\end{tabular} & \multicolumn{1}{l}{} & \begin{tabular}[c]{@{}c@{}}Without\\ navigation\end{tabular} & \begin{tabular}[c]{@{}c@{}}With\\ navigation\end{tabular} \\ \midrule
Number of objects        & 14        & 46        &           & 52.9               & 92.2                                                        &                      & 62.3                                                           &  94.7                                                        \\
Number of kills          & 167       & 138       &           & 43.0               & 66.8                                                         &                      & 32.0                                                            & 43.0                                                         \\
Number of deaths         & 36        & 25        &           & 15.2               & 14.6                                                         &                      & 10.0                                                            & 6.0                                                         \\
Number of suicides       & 15        & 10        &           & 1.7                & 3.1                                                         &                      & 0.3                                                            & 1.3                                                         \\
Kill to Death Ratio      & 4.64      & 5.52      &           & 2.83               & 4.58                                                         &                      & 3.12                                                            & 6.94                                                         \\ \bottomrule
\end{tabular}
\caption{Performance of the agent against in-built game bots with and without navigation. The agent was evaluated 15 minutes on each map. The performance on the full deathmatch task was averaged over 10 train maps and 3 test maps.}
\label{tab:splitting}
\end{table*}

\subsection{Hyperparameters}
All networks were trained using the RMSProp algorithm and minibatches of size $32$. Network weights were updated every $4$ steps, so experiences are sampled on average 8 times during the training \cite{van2015deep}. The replay memory contained the one million most recent frames. The discount factor was set to $\gamma = 0.99$. We used an $\epsilon$-greedy policy during the training, where $\epsilon$ was linearly decreased from $1$ to $0.1$ over the first million steps, and then fixed to $0.1$.

Different screen resolutions of the game can lead to a different field of view. In particular, a 4/3 resolution provides a 90 degree field of view, while a 16/9 resolution in Doom has a 108 degree field of view (as presented in Figure~\ref{fig:scrn}). In order to maximize the agent game awareness, we used a 16/9 resolution of 440x225 which we resized to 108x60. Although faster, our model obtained a lower performance using grayscale images, so we decided to use colors in all experiments.

\subsection{Scenario}
We use the ViZDoom platform \cite{kempka2016vizdoom} to conduct all our experiments and evaluate our methods on the deathmatch scenario. In this scenario, the agent plays against built-in Doom bots, and the final score is the number of frags, i.e. number of bots killed by the agent minus the number of suicides committed. We consider two variations of this scenario, adapted from the ViZDoom AI Competition:
\subsubsection{Limited deathmatch on a known map.}
The agent is trained and evaluated on the same map, and the only available weapon is a rocket launcher. Agents can gather health packs and ammo.
\subsubsection{Full deathmatch on unknown maps.}
The agent is trained and tested on different maps. The agent starts with a pistol, but can pick up different weapons around the map, as well as gather health packs and ammo. We use 10 maps for training and 3 maps for testing. We further randomize the textures of the maps during the training, as it improved the generalizability of the model.
\\
The limited deathmatch task is ideal for demonstrating the model design effectiveness and to chose hyperparameters, as the training time is significantly lower than on the full deathmatch task. In order to demonstrate the generalizability of our model, we use the full deathmatch task to show that our model also works effectively on unknown maps.

\subsection{Evaluation Metrics}
For evaluation in deathmatch scenarios, we use Kill to death (K/D) ratio as the scoring metric. Since K/D ratio is susceptible to ``camper'' behavior to minimize deaths, we also report number of kills to determine if the agent is able to explore the map to find enemies. In addition to these, we also report the total number of objects gathered, the total number of deaths and total number of suicides (to analyze the effects of different design choices). Suicides are caused when the agent shoots too close to itself, with a weapon having blast radius like rocket launcher. Since suicides are counted in deaths, they provide a good way for penalizing K/D score when the agent is shooting arbitrarily. 

\subsection{Results \& Analysis}
\label{sec:results}
Demonstrations of navigation and deathmatch on known and unknown maps are available \href{https://www.youtube.com/playlist?list=PLduGZax9wmiHg-XPFSgqGg8PEAV51q1FT}{here}\footnote{\url{https://www.youtube.com/playlist?list=PLduGZax9wmiHg-XPFSgqGg8PEAV51q1FT}}. Arnold, an agent trained using the proposed Action-Navigation architecture placed second in both the tracks Visual Doom AI Competition with the highest K/D Ratio \cite{chaplot2016arnold}.

\subsubsection{Navigation network enhancement.} Scores on both the tasks with and without navigation are presented in Table~\ref{tab:splitting}. The agent was evaluated 15 minutes on all the maps, and the results have been averaged for the full deathmatch maps. In both scenarios, the total number of objects picked up dramatically increases with navigation, as well as the K/D ratio. In the full deathmatch, the agent starts with a pistol, with which it is relatively difficult to kill enemies. Therefore, picking up weapons and ammo is much more important in the full deathmatch, which explains the larger improvement in K/D ratio in this scenario. The improvement in limited deathmatch scenario is limited because the map was relatively small, and since there were many bots, navigating was not crucial to find other agents. However, the agent was able to pick up more than three times as many objects, such as health packs and ammo, with navigation. Being able to heal itself regularly, the agent decreased its number of deaths and improved its K/D ratio. Note that the scores across the two different tasks are not comparable due to difference in map sizes and number of objects between the different maps. The performance on the test maps is better than on the training maps, which is not necessarily surprising given that the maps all look very different. In particular, the test maps contain less stairs and differences in level, that are usually difficult for the network to handle since we did not train it to look up and down.

\subsubsection{Comparison to human players.}
Table~\ref{tab:results} shows that our agent outperforms human players in both the single player and multiplayer scenarios. In the single player scenario, human players and the agent play separately against 10 bots on the limited deathmatch map, for three minutes. In the multiplayer scenario, human players and the agent play against each other on the same map, for five minutes. Human scores are averaged over 20 human players in both scenarios. Note that the suicide rate of humans is particularly high indicating that it is difficult for humans to aim accurately in a limited reaction time.

\subsubsection{Game features.} Detecting enemies is critical to our agent's performance, but it is not a trivial task as enemies can appear at various distances, from different angles and in different environments. Including game features while training resulted in a significant improvement in the performance of the model, as shown in Figure~\ref{fig:graphs}. After 65 hours of training, the best K/D score of the network without game features is less than $2.0$, while the network with game features is able to achieve a maximum score over $4.0$.

Another advantage of using game features is that it gives immediate feedback about the quality of the features given by the convolutional network. If the enemy detection accuracy is very low, the LSTM will not receive relevant information about the presence of enemies in the frame, and Q-learning network will struggle to learn a good policy. The enemy detection accuracy takes few hours to converge while training the whole model takes up to a week. Since the enemy detection accuracy correlates with the final model performance, our architecture allows us to quickly tune our hyperparameters without training the complete model.

For instance, the enemy detection accuracy with and without dropout quickly converged to 90\% and 70\% respectively, which allowed us to infer that dropout is crucial for the effective performance of the model. Figure~\ref{fig:graphs} supports our inference that using a dropout layer significantly improves the performance of the action network on the limited deathmatch.

As explained in Section~\ref{sec:game-feature-augmentation}, game features surprisingly don't improve the results when used as input to the DQN, but only when used for co-training. This suggests that co-training might be useful in any DQN application even with independent image classification tasks like CIFAR100.

\section{Related Work}

\newcite{mcpartland2008learning} and \newcite{tastan2011learning} divide the tasks of navigation and combat in FPS Games and present reinforcement learning approaches using game-engine information.
\newcite{koutnik2013evolving} previously applied a Recurrent Neural Network to learn TORCS, a racing video game, from raw pixels only. 
\newcite{kempka2016vizdoom} previously applied a vanilla DQN to simpler scenarios within Doom and provide an empirical study of the effect of changing number of skipped frames during training and testing on the performance of a DQN.

\section{Conclusion}
\label{sec:conclusions}
In this paper, we have presented a complete architecture for playing deathmatch scenarios in FPS games. We introduced a method to augment a DRQN model with high-level game information, and modularized our architecture to incorporate independent networks responsible for different phases of the game. These methods lead to dramatic improvements over the standard DRQN model when applied to complicated tasks like a deathmatch. We showed that the proposed model is able to outperform built-in bots as well as human players and demonstrated the generalizability of our model to unknown maps. Moreover, our methods are complementary to recent improvements in DQN, and could easily be combined with dueling architectures \cite{wang2015dueling}, and prioritized replay \cite{schaul2015prioritized}.

\section{Acknowledgements}
We would like to acknowledge Sandeep Subramanian and Kanthashree Mysore Sathyendra for their valuable comments and suggestions. We thank students from Carnegie Mellon University for useful feedback and for helping us in testing our system. Finally, we thank the ZDoom community for their help in utilizing the Doom game engine.

\bibliographystyle{aaai}
\bibliography{references}

\end{document}